\documentclass[a4,conference]{IEEEtran}

\newcommand{\etal}{{\it et~al.}}

\usepackage{amsmath, scalerel}

\usepackage[linesnumbered,ruled]{algorithm2e}
\usepackage{graphicx}
\graphicspath{ {images/} }
\usepackage{pgfplots, lipsum}
\usepackage{textcomp}
\pgfplotsset{compat = newest}

\pgfplotsset{width=10cm,compat=1.9}
\usepgfplotslibrary{external}
\usepackage{tikz}
\usetikzlibrary{arrows}
\usepackage{caption}
\usepackage{siunitx}
\usepackage[font={small}]{caption}
\usepackage{amsmath}
\usepackage[export]{adjustbox}
\usepackage{lipsum}
\usepackage{subcaption}

\usepackage{xcolor}
\usepackage{pgfplots}
\usepackage{tikz}
\usepackage{pgfkeys}
\usepackage{url}
\usepackage{pgfplotstable}
\usepackage{subcaption}
\usepackage{filecontents}
\usepackage{multirow}
\usepackage{pgf}

\usepackage{pgfplots}
\usepackage{pgfplotstable}
\pgfplotsset{compat = newest}
\usepackage{balance}

\makeatletter
\def\BState{\State\hskip-\ALG@thistlm}

\makeatother

\usepackage{cite}

\ifCLASSINFOpdf
  
\else
  
\fi

\begin{document}

\title{Machine Learning Technique  Predicting  Video Streaming Views to Reduce Cost of Cloud Services  
}

\author{\IEEEauthorblockN{Mahmoud Darwich}\\ 
\IEEEauthorblockA{ Department of Mathematics and Computer Science\\
University of Mount Union, Alliance, Ohio 44601\\
Email:darwicma@mountunion.edu}
}
\author{\IEEEauthorblockN{ Mahmoud Darwich}
\IEEEauthorblockA{\textit{Department of Mathematics and Computer Science} \\
\textit{University of Mount Union}\\
Alliance, Ohio 44601, USA \\
darwicma@mountunion.edu}
}

\maketitle

\begin{abstract}

Video streams tremendously occupied the highest portion of the online traffic. Multiple versions of a video are created to fit the user's device specifications. In cloud storage, Keeping all versions of frequently accessed video streams in the repository for the long term imposes a significant cost paid by video streaming providers. Generally, the popularity of a video changes each period time, which means the number of views received by a video could be dropped, thus, the video must be deleted from the repository. Therefore, in this paper, we develop a method that predicts the popularity of each video stream in the repository in the next period. On the other hand, we propose an algorithm that utilizes the predicted popularity of a video to compute the storage cost, and then it decides whether the video will be kept or deleted from the cloud repository. The experiment results show a cost reduction of the cloud services by 15\% compared to keeping all video streams.

\end{abstract}


\IEEEpeerreviewmaketitle

\section{Introduction }

In the recent decade, the technology of filming any social events, places, individuals, and objects is easily accessible by any holder of a smartphone or other devices films. Moreover, the extensive spread out of social media platforms increases the number of video streams over the internet significantly. Additionally, every device plays the version of a video stream that fits its characteristics, Therefore, a lot of versions of the same video stream are made ready to be displayed on all different types of devices. Thus, this increases the video streams incredibly across the internet platform. Designing storage for all video streams becomes remarkably challenging. Thankfully, the cloud was revealed as a technology to tackle the problem of building physical database centers to fit all the huge data \cite{xiangbo2020} \cite{darwich22}.

Cloud was adopted strongly by tech companies to ease the operation of processing and securing data. Therefore, cloud services are available for small and enterprise businesses with affordable prices and high benefits. Cloud offers virtual machines (VM) at different processors speed and utilization in addition to the storage option, which is rated on an hourly and monthly basis respectively
\cite{amazon}.

Transcoding is an operation to convert a video stream into different versions. However, transcoding is an extensive process that requires powerful virtual machines. Accordingly, high costs are imposed by using VMs in the cloud. Alternatively, storing servers are less expensive and used to store the pre-made (i.e. pre-transcoded) versions of video streams e-transcoded video. Generally, storing option in the cloud is applied for frequently accessed video streams (FAVs).

Amazon Web Services (AWS) is one of the reputed cloud operators and we use their standards to run the experiments. Our method could apply to any cloud provider standards. AWS rates are displayed as follows:
\begin{itemize}
\item \texttt{S3 Standard}: It is typical storage in the cloud for all contents.
\item \texttt{Amazon Elastic Compute Cloud (Amazon EC2)}: presents the VMs that offer the computational services and it comes under different types.
.
\end{itemize}

We define the research problem in this paper to predict the views of the currently stored video streams in the repository in the next period. Estimating the views in the next time cycle could reduce the storage cost of video streams in the cloud.

The rest of the paper is organized as follows: related works are presented in section II. The proposed popularity prediction is discussed in Section III. The experiments are implemented in section IV. Section V shows the results and then section VI presents the conclusion.

\section{Related Work}

Darwich \etal \cite{darwichhotness} developed approaches to reduce the cost of cloud services by organizing the video streams in the repository. Particularly, they developed an algorithm that computes the hotness of videos and they apply this criterion to manage the videos. Their results show a reduction in cloud services costs.

On the other hand, Darwich \etal \cite{darwich2016} presented a method that filters the videos to be stored or deleted and then re-transcoded upon request. They applied the cost ratio in their algorithm to decide on either storing or transcoding. Their proposal shows a reduction in cloud costs.

Darwich \etal \cite{darwich2020} designed a scheme that distributes the videos into different types of cloud storage. Their design is based on video clustering. Their experimental results show a reduction in cloud cost.

Jokhio \etal \cite{jokhio} presented an approach to reduce the cloud cost by measuring the popularity of each video stream in the repository. they applied a weighted graph to select either transcoding or storing.

Gao \etal \cite{gao} proposed a method that split videos into segments and then applies the view distribution to reduce the cost of cloud services.

Zhao \etal \cite{zhao} proposed a scheme to build a relationship between the video version and its views. They developed a strategy to form a relationship between video versions. their results show a reduction in the cost of cloud services.

\section{Proposed Popularity Prediction}
\subsection{Popularity}
The popularity $\rho_i$ of a video stream $i$ is measured as the number of views received by the video in a period. In this research, We consider counting the views of each video in the repository on an hourly basis for 30 days. The popularity plays a key role in the cost of cloud services for video streams in the repository.

\subsection{Storage Cost}
The Storage cost of a video stream $i$ in the cloud is dependent on the video size and the storage price rated in \$ gigabytes per hour. The storage cost is calculated as shown in Equation \ref{storage}
\begin{equation}
C_{si}=\dfrac{P_s*\sigma_i}{2^{10}}
\end{equation}\label{storage}
Where $P_s$ is the cloud storage cost and $\sigma_i$ is the size of video $i$ in megabytes

\subsection{Transcoding Cost}
The transcoding cost of a video stream $i$ in the cloud is dependent on the number of views received by the video and the price of a virtual machine in the cloud, which is rated in \$ per hour. Assuming video $i$ in the cloud repository, every time the video is viewed, the video is transcoded to a version to fit the viewer's device specifications. the transcoding cost of the video received views $\nu$ is computed as illustrated in the Equation \ref{trans}
\begin{equation}
C_{ti}=\dfrac{P_t*\nu_i}{3600}
\end{equation} \label{trans}
where $P_t$ is the price of VM rated on an hourly basis.

\subsection{Cost Ratio}\label{cr}
The cost ratio $\psi_i$ of a video stream $i$ is computed as the storage cost of the video stream in the repository divided by its transcoding cost using the Equation \ref{storage} and \ref{trans}.
\begin{equation}
\psi_i=\dfrac{C_{si}}{C_{ti}}
\end{equation}\label{ratio}
When the cost ratio is greater than or equal to 1, that means it is worth storing the video version. When the cost ratio is less than 1, the video version must be deleted and transcoded upon request.
The following conditions of $\psi_i$ are detailed as follows:
\begin{equation}
\psi_i \colon
\begin{cases}
\text {if} \, \psi_i>1 \rightarrow \text{transcode video}\\
\text {if} \, \psi_i <= 1 \rightarrow \text{store video}
\end{cases}
\end{equation}

\subsection{Proposed Prediction Method}
In section \ref{cr}, we calculated the cost ratio, if it is greater or equal to 1, the video version should be stored for a period of time. Since the fees of cloud storage are charged monthly. Thus, we assume the period time of storing any video version is one month only. Therefore, it is essential to figure out if the stored video in the current month will be kept in the storage for the next month. The clue, which determines whether the stored will be kept in the repository or not during the next period of time, is the popularity (i.e. number of views) of the video. Predicting the number of views of each stored video in the next period cycle is the key to deciding
if it will be kept or not. In this paper, we apply a machine learning technique to predict the popularity of each stored video in the repository. Particularly, we implement the method of linear regression on the views of the current video streams in the repository.

\subsection{ Proposed algorithm }

{\SetAlgoNoLine%

\begin{algorithm}

 \SetKwInOut{Input}{Input}
 \SetKwInOut{Output}{Output}

 \Input{Frequently accessed video $i $\\
 	 Size of each frequently accessed video$\sigma_i $ \\
	 Cloud Storages price: $P_s$\\
	Transcoding price: $P_t$\\
	 Number of views of each : $ \nu_i$\\

 }
 \Output{Cost of streaming videos in cloud\\ }

Apply Linear Regression on the views of the videos and predict the views: $ \nu_{i-predict}$ \\
 \For { each video $i$}{
Storage Cost:$ C_{si} \leftarrow \dfrac{P_s*\sigma_i}{2^{10}}$ \\
 Estimated Transcoding Cost: $C_{ti-estimated} \leftarrow \dfrac{P_t*\nu_{i-predict}}{3600}$ \\
 Estimaed Cost Ratio: $\psi_{i-estimated} \leftarrow \dfrac{C_{si}}{C_{ti-estimated}}$\\
\eIf{ $\psi_{i-estimated}>=1$ }{keep the video}{delete video}

}

Totat Cost:$ \leftarrow \dfrac {\sum P_s*\sigma_i}{2^{10}} +\dfrac{ \sum C_{si}}{C_{ti-estimated}}$

 \caption{GOPs storing cost}
 \label{al2}
  
\end{algorithm}
}

 The algorithm's goal is to predict the views of each video in the next period and recalculates the transcoding cost and ratio cost. if the ratio comes to less than 1, that means the cost of storing is less than the cost of transcoding and vice versa. The inputs to the algorithm are, video size and its views, storage and transcoding cost. The algorithm output is to predict the views for each video and recalculates the cost of cloud services. In step 1, we apply the machine learning technique by implementing the linear regression method on the video stream views during the current period which we consider as  30 days. The views for each video are predicted for the next period of time. In step 2, we compute the storage cost for each video.  In step 3, we recalculate the transcoding cost for each video using the predicted views. In step 4, we measure the cost ratio, if it is less or equal to one, then we keep the video. If the cost ratio is greater than one we delete the video and re-transcode upon request as shown in steps 5 and 6 respectively. In the last step, the algorithm recalculates the total cost of cloud services.

\balance

\section{Experiment Setup }

\subsection{Videos Synthesis}
As it is difficult to gain an access to video stream repositories of the providers. Therefore, we came up with a way to synthesize the video streams by implementing the techniques in \cite{darwich2016}. Particularly, we synthesized the useful information of videos like the video size and its transcoding time which are used in the experiments.

\subsection{ Amazon Storage Rates}

Table \ref{tab:storageprice} displays the AWS rates in US dollars

 \begin{table}[ht]
\centering 
\caption{AWS  S3 Standard storage and EC2 rates in USD }
\begin{tabular}{c c c c} 
\hline\hline 
 S3 Standard  Storage & EC2 \\ [0.5ex] 
\hline 

\texttt{ \$0.05 GB/hour} &\texttt{ \$0.023 GB/month} \\

\hline
\hline 
\end{tabular}

\label{tab:storageprice}
\end{table}

\subsection{ Approaches for Comparison}

We evaluate our proposed method by implementing related methods and comparing them.
\begin {itemize}

\item \emph{Partial storing} method in \cite{darwich2016} , it stores a part of video stream which receives frequent accesses in the cloud standard storage \texttt{S3 Standard}.
\item \emph{Clustering video streams} method in \cite{darwich2020}. It clusters the frequently accessed video stream and stores them in different storages in the cloud
\item \emph{Clustering GOPs Clustering} method in \cite{darwich2020}. It clusters the frequently accessed GOPs from different videos and stores them in different storages in the cloud

\end{itemize}

\section { Simulation Results}


\begin{figure}
\centering

\resizebox{1\columnwidth}{!}{%
\begin{tikzpicture} 
 \begin{axis}[
 width =0.51*\textwidth,
 height = 7cm,
 ymajorgrids=true,
 legend style={font=\fontsize{7}{5}\selectfont},
 ylabel = {Total cost (\$)},
 xlabel={Percentage of frequently accessed videos in   the repository},
 xticklabel={$\pgfmathprintnumber{\tick}\%$},
 xtick = data,
 legend pos= south west,
 ]

 \addplot
 coordinates {(5, 694)
 (10, 662)
 (15, 643)
 (20, 613)
 (25, 566)
 (30, 533)};

 \addplot
 coordinates {(5, 650)
 (10, 620)
 (15, 600)
 (20, 560)
 (25, 530)
 (30, 480)};

\addplot+[mark=triangle]
 coordinates {(5, 600)
 (10, 590)
 (15, 560)
 (20, 520)
 (25, 470)
 (30, 390)};

\addplot
 coordinates {(5, 590)
 (10, 580)
 (15, 550)
 (20, 500)
 (25, 440)
 (30, 370)};
  
 \legend{ Partially storing approach \cite{darwich2016}, Video Clustering approach \cite{darwich2020},  GOPs Clustering approach \cite{darwich21}, Proposed Popularity Prediction}

 \end{axis}
  
\end{tikzpicture}


}
\caption{Cost comparison of the four approaches Partially storing approach \cite{darwich2016}, Video Clustering approach \cite{darwich2020},  GOPs Clustering approach\cite{darwich21}, Proposed Popularity Prediction when the number of frequently accessed videos varies }
 \label {favs}
\end{figure}
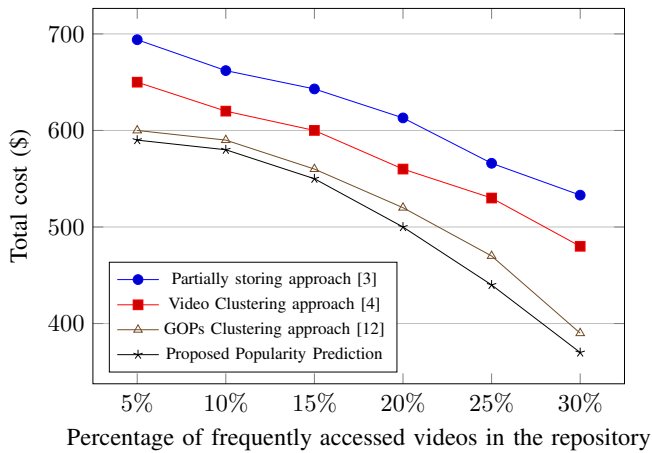

\begin{figure*}
\centering

\resizebox{0.7\textwidth}{!}{%
\begin{tikzpicture}
\begin{axis}[
    xmin = 0, xmax =720,
    ymin = 0, ymax = 1000,
width =0.71*\textwidth,
 height = 8cm,
    minor tick num = 1,
    major grid style = {lightgray},
    minor grid style = {lightgray!25},
 legend style={font=\fontsize{8}{5}\selectfont},
    xlabel = {Hour ($h$)},
    ylabel = {Number of Views ($v$)},
    legend cell align = {left},
    legend pos = north west
]
 
\addplot[
    teal, 
    only marks
] table[x = h, y = v] {views.dat};
 
\addplot[
    thick,
    orange
] table[
    x = h,
    y = {create col/linear regression={y=v}}
] {views.dat};
 
\addlegendentry{Data}
\addlegendentry{
    Linear regression: $ v =
    \pgfmathprintnumber{\pgfplotstableregressiona}
    \cdot h
    \pgfmathprintnumber[print sign]{\pgfplotstableregressionb}$
};
 
\end{axis}

\end{tikzpicture}
 
	}
\caption{  Linear Regression technique on the hourly views of video streams. the horizontal x axis represents the hours of one month (24*30)=720. The vertical y-axis represents the number of views in each hour during the 30 days }
\label{regression}
\end{figure*}
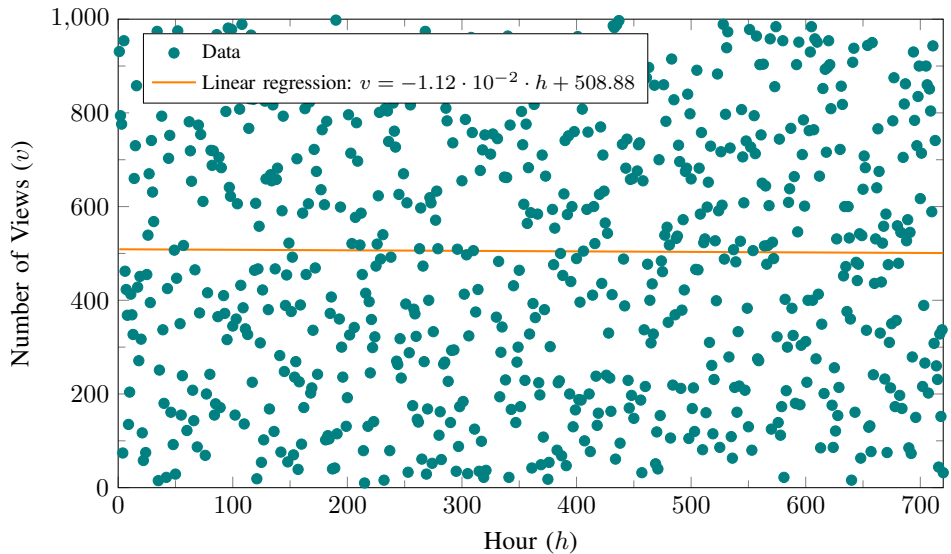

 \begin{table*}[ht]
\centering 
\caption{ Cost comparison in USD when the percentage of frequently accessed video changes }
\begin{tabular}{c c c c c c c } 
\hline\hline 
 FAVs \% & Fully Transconding & Full Storing& Partial Storing \cite{darwich2016}& Videos Clustering \cite{darwich2020} &  GOPs Clustering \cite{darwich21} & Proposed Popularity Prediction \\ [0.5ex] 
\hline 

5\% &1598& 839 & 694 & 	 650 & 600	&590	 \\
10 \% &1598& 	842	&662 &	620 &	590 &580	\\                                         
15 \% &1598&	863	& 643 &	600 &	560 &550 	\\
20 \% &1598&	947	&613 &	560 &	520 & 500	\\
25 \% &1598& 1424 & 566 & 530 & 470& 440  \\
30 \% &1598& 3137& 533 &  480 & 390 & 370\\
\hline
\hline 
\end{tabular}
\label{tab:cost}
\end{table*}

Figure \ref{favs} compares the four approaches. The horizontal x-axis represents repositories of video streams with different percentages of frequently accessed videos. The vertical y-axis represents the total cost of cloud services including the sum of both transcoding and storing costs. As we notice from the figure generally the cost of the four approaches is high. The reason behind that is the repository contains a small percentage of frequently accessed videos, which means more transcoding operations are implemented. Thus the total cost comes high.
As the number percentage of frequently accessed videos increases, the cost of all approaches decreases because fewer transcoding operations are executed. When 30\% of the repository is for frequently accessed videos, the proposed approach outperforms all other approaches because there are videos that were considered highly accessed during the current period. While in the next period, these videos are no longer highly accessed and thus they are deleted from the repository. Therefore, the total cost is decreased remarkably.
The proposed approach reduces the cost by \%15 compared to the GOPs clustering approach.

Table \ref{tab:cost} displays the cost numbers for two approaches( fully storing and fully transcoding) which we didn't include in Figure \ref{favs} and the four approaches. We eliminated the first two approaches as they cost highly.

Figure \ref{regression} represents the implementation of the linear regression on vies of video streams during a period to predict the views in the next period. The data shows the views of each video on an hourly basis. That means in 30 days, there are 720 hours. The horizontal x-axis represents the 720 hours and the vertical y-axis represents the views of each video
the linear equation is displayed in the figure.

\section{Conclusion}
In this research, we proposed an approach to predict the number of views and it recalculates the cost of cloud services for the next period. Our approach shows an efficient way to reduce the cost by implementing machine learning on the video stream views. Moreover, the proposed approach shows a significant reduction when the repository contains a high percentage of frequently accessed video streams.





%

\vfill\eject
\balance


\end{document}